\documentclass[11pt]{article}
\usepackage{mtsummit2019}
\usepackage{times}
\usepackage{url}
\usepackage{latexsym}
\usepackage[small,bf]{caption} 
\usepackage[english]{babel}
\usepackage[utf8x]{inputenc}

\usepackage{amsmath}
\usepackage{graphicx}
\usepackage[numbers]{natbib}

\usepackage{float}
\usepackage[justification=justified]{caption}
\usepackage{enumitem} 
\usepackage{dirtytalk}
\usepackage{array}
\usepackage{multirow}

\usepackage{siunitx}
\usepackage{color}

\renewcommand{\citet}{\cite}

\title{Transductive Data-Selection Algorithms for Fine-Tuning Neural Machine Translation}

\author{Alberto Poncelas \and Gideon Maillette de Buy Wenniger \and Andy Way\\
ADAPT Centre, School of Computing, \\ Dublin City University, Dublin, Ireland\\ {\tt \{firstname.lastname\}@adaptcentre.ie}
}

\begin{document}

\maketitle

\begin{abstract}
Machine Translation models are trained to translate a variety of documents from one language into another. However, models specifically trained for a particular characteristics of the documents tend to perform better. Fine-tuning is a technique for adapting an NMT model to some domain. In this work, we want to use this technique to adapt the model to a given test set. In particular, we are using transductive data selection algorithms which take advantage the information of the test set to retrieve sentences from a larger parallel set. 

In cases where the model is available at translation time (when the test set is provided), it can be adapted with a small subset of data, thereby achieving better performance than a generic model or a domain-adapted model.
\end{abstract}

\section{Introduction}

Machine Translation (MT) models aim to generate a text in the target language which corresponds to the translation of a text in the source language, the test set. These models are trained with a set of parallel sentences so they can learn how to generalize and infer a translation when a new document is seen.

In the field of MT, Neural Machine Translation (NMT) models tend to achieve the best performances when large amounts of parallel sentences are used. However, relevant data is more useful than having more data. Previous studies \citep{silva2018extracting} showed that models trained with in-domain sentences perform better than general-domain models.

However, training models for domains that are distant from general domains, such as scientific documents, is not always a simple task as parallel sentences are not always available. In addition, identifying the domain adds complexity if the domain of the document to be translated is too specific. The alternative explored in this work is to build models adapted to a given test set.

In order to build task-specific models, data selection algorithms play an important role as they retrieve sentences from the training data. Data selection methods can be classified \citep{eetemadi2015survey} according to the criteria considered to select sentences (e.g. select sentences of a particular domain, good quality sentences, etc.). In this work, we use the transductive \citep{Vapnik1998} data selection methods which use the document to be translated to select sentences that are the most relevant for translating such text. 

In some cases, the organizations in charge of translating a document are also the owner of the translation model and training data. Therefore, knowing the test set is an advantage that can be helpful for adapting the generic MT model towards the test set \citep{utiyama2009two,liu2012locally}.

The approaches presented here consist of building a single NMT model and delay part of the process of training data for adapting the model when the test set is available. Although this implies increasing the time involved in translating a document, it also has some benefits.

First, using a single model causes storing multiple task-adapted models not to be necessary. Moreover, identifying the domain of the document (and so, the most appropriate model) before the translation is also avoided. In addition, due to the fine-grained adaptation, other characteristics that may have not been foreseen (e.g. formal or informal register, technical or literal vocabulary, the gender of the speaker etc.) are also considered.




This paper presents the performance of three transductive data selection algorithms (TA), applied to NMT models, showing how these models can be improved by adapting them with a small set of data. The TAs are executed using the test set as seed, but there are other approaches such as using an approximated target-side \citep{poncelas2018data,poncelas2018adapt}.


The remainder of this paper is structured as follows. In Section \ref{sec:finetun_RQ}, we state the research questions that we want to investigate. Section \ref{sec:related_work} contains some insights of other works that are related to this and Section \ref{sec:transductive_data_selection_algorithms} describes the data selection methods used in the experiments.  In Section \ref{sec:finetun_analysis} we perform an analysis of fine-tuning and in Section \ref{sec:baseline_results} we build the models used as baselines in later experiments. The results of the main experiments are explained in Section \ref{sec:experiments} and finally, in Section \ref{sec:conclusions}, we conclude and indicate further research that can be carried out in the future.

\section{Research Questions}
\label{sec:finetun_RQ}


In this work, we are using a general-domain data set to build an NMT model. Then, this model will be adapted, performing fine-tuning, to two different test sets in two domains: news and health. The data used to adapt the model is retrieved by the algorithms described in Section \ref{sec:transductive_data_selection_algorithms}. These methods will retrieve sentences from: (i) the general domain data; (ii) different in-domain datasets; and (iii) from a concatenation of both the general domain and in-domain set. Therefore the research questions we propose to explore are the following three:

\begin{enumerate}
    \item Can a model fine-tuned with a subset of data outperform the model trained with general domain data?\\
    
    The work of \citet{poncelas2018feature} showed that performing fine-tuning on a subset of data (used to build the model) yields small improvements (and not statistically significant at level p=0.01). A limitation in their experiments is that, as BPE is not applied, the vocabulary of the adapted model remains the same as the general model. As in these experiments we are processing the data using BPE, the limitation of the vocabulary should disappear (as sub-words are considered rather than complete words). We are interested in exploring whether performing fine-tuning with a subset of the data (in which BPE was applied) can improve the base model.
    
    \item Can a model fine-tuned with a subset of in-domain data outperform the model fine-tuned with the complete data set?\\
    
    The general uses of fine-tuning \citep{luong2015stanford,freitag2016fast} consist of using in-domain data set to adapt a model. However, we want to investigate whether applying data selection in smaller \textit{in-domain} set can also lead to improvements.
    
    \item Can a model fine-tuned with a dataset mixture of general-domain and in-domain data outperform the previous-mentioned models?\\
    
    By considering both datasets (general and in-domain data), the number of candidate sentences is increased. This also poses a challenge to the transductive algorithm as most of the candidate sentences are not in-domain. We are interested in exploring whether these algorithms can successfully retrieve sentences that lead to improvements.

\end{enumerate}

\section{Related Work}
\label{sec:related_work}

There are several adaptation techniques for NMT. \citet{chu2018survey} structure them into two main groups, \textit{data centric} (techniques which involve augmenting or modifying the training data) and \textit{model centric} (techniques which involve modifying the architecture or the procedure with which the model is trained). In this paper, we use a combination of both as we use data selection methods (data centric) and fine-tuning (model centric).

The technique of fine-tuning \citep{luong2015stanford,freitag2016fast} consists of training an NMT model with a general domain data set until convergence, and then using an in-domain set for the last epochs.

The work of \citet{van2017dynamic} showed that training an NMT model using less (but more in-domain) data each epoch achieves improvements over a model trained with all data. Their experiments include weighting the sentences using Cross Entropy Difference \citep{axelrod2011domain}, and then, each epoch $e$ the top-$N_e$ sentences are used as training data where $N_1 \geq N_e\geq N_{last}$


A proposal in which they use the test set to adapt the model is the work of \citet{li2018one}. In particular, they fine-tune a pre-built NMT model for each sentence in the test set. They use three methods to retrieve the sentences that are the most similar to a sentence of the test set: (i) Levenshtein distance \citep{levenshtein1966binary}; (ii) cosine similarity of the average of the word embeddings \citep{mikolov2013distributed}; and (iii) the cosine similarity between hidden states of the encoder in NMT. The main difference with our work is that they adapt the model sentence-wise (one model for each sentence) whereas the adaptations presented here are document-wise (one model for each test set). Although performing adaptations sentence-wise gives more fine-grained adaptations, it also has several disadvantages: (i) the computational cost is higher as there are several iterations (as many as sentences in the test set) of selecting data and fine-tuning; (ii) the usage of the data is less efficient as a same sentence can be extracted multiple times (in different iterations); and (iii) using different models for each sentence has the potential risk of performing translations that are not consistent throughout the entire document.


\section{Transductive Data Selection Algorithms}
\label{sec:transductive_data_selection_algorithms}

In this work, we investigate data selection methods that exploit the information of the test set to retrieve sentences. These methods select a subset of from the parallel set $(S,T)$ used as training data. In particular, they select sentences based on overlaps of {\em n}-grams between the test set $S_{test}$ and the source side of the parallel data $S$. In this work, we explore the following three techniques:

\paragraph{TF-IDF Distance Method:}
Distance methods measure how close two sentences are by using metrics as Levenshtein distance (which computes the minimum number of insertion, deletions or substitutions of characters that are necessary to transform one sentence into the other) to score the similarities. \citet{hildebrand2005adaptation} propose \textit{TF-IDF distance} i.e. to use cosine between TF-IDF \citep{tfidf1973} vectors as distance metric. In their work, for each $s_{test}\in S_{test}$ the top sentences from $S$ are selected. Although they are aware that the resulting set contains duplicated sentences, in their experiments the models containing duplicated sentences achieve slightly better results.



TF-IDF measures the importance of the terms in a set of documents. Each document $D$ can be represented as a vector of terms $\mathbf{w}_D=(w_{1},w_{2}, \dots w_{|V|})$, where $|V|$ is the size of the vocabulary. Each $w_{k}$ is calculated as in \eqref{eq:tfidf}:
\begin{equation}\label{eq:tfidf}
w_{k}=tf_{k}*log(idf_k)
\end{equation}

\noindent where $tf_{k}$ is the term frequency (TF) of the k-th term in $D$, i.e. the number of occurrences, and $idf_k$  is the inverse document (IDF) frequency of the k-th term, as in \eqref{eq:idf}:

\begin{equation}\label{eq:idf}
idf_k=\frac{\#documents}{\# documents \; containing \;  term \; k}
\end{equation}

The similarity between two sentences $a$ and $b$ is computed as the inverse of the cosine distance of their TF-IDF vectors, $\mathbf{w}_a$ and $\mathbf{w}_b$, as in Equation~\eqref{eq:tfidf_sim}:

\begin{equation}\label{eq:tfidf_sim}
sim(a,b)=1-cos(\mathbf{w}_a,\mathbf{w}_b)\\=1-\frac{\mathbf{w}_a \cdot \mathbf{w}_b}{|\mathbf{w}_a| |\mathbf{w}_b|}
\end{equation}

In the TFIDF transductive method, each sentence $s$ in the \textit{Candidate data} $S$ is scored according to the highest similarity with a sentence $r$ from the test set $S_{test}$ computed as in Equation~\eqref{eq:score_tfidf}:

\begin{equation}\label{eq:score_tfidf}
score(s)=\max_{r \in S_{test}} sim(s,r)
\end{equation}


\paragraph{Infrequent {\em n}-gram Recovery} (INR):
\citet{parcheta2018data} propose extracting those sentences containing {\em n}-grams from the test set that are considered infrequent \citep{gasco2012does} (so frequent words such as stop words are ignored).

A sentence $s$ is scored according to the number of infrequent {\em n}-grams shared with the set of sentences of the test set $S_{test}$. It is computed as in Equation~\eqref{eq:infreq_ngr_recover}:


\begin{equation}\label{eq:infreq_ngr_recover}
score(s)=\sum_{ngr \in \{ S_{test} \bigcap  s \} }  max(0,t-C_L(ngr)) 
\end{equation}


\noindent where $C_L(ngr)$ is the count of $ngr$ in the selected set of sentences $L$ (those that have been selected already). $t$ is the number of occurrences of an {\em n}-gram to be considered infrequent. If the number of occurrences of $ngr$ is above the threshold $t$ then $ngr$ is considered frequent {\em n}-gram (the component $max(0,t-C_S(ngr))$ is 0) and it does not contribute for scoring the sentence. When a sentence is added to the selected pool the count of the {\em n}-gram in the candidate data $C_L(ngr)$ is updated \citep{gasco2012does}.

\paragraph{Feature Decay Algorithms} (FDA):
Feature Decay Algorithms \citet{biccici2011instance} selects data trying to maximize the variability of {\em n}-grams in the selected data by decreasing their value as they are added to a  selected pool $L$, which eventually becomes the selected data. 

In order to do that, the {\em n}-grams in the test set are extracted and assigned an initial value. Each sentence in the set of candidate sentences has an importance score (i.e. the normalized sum of the score of its {\em n}-grams) of being selected.

Then, iteratively, the sentence with the highest score in the candidate data is selected and added to a set of selected pool $L$. In addition, the values of the {\em n}-grams of the selected sentence are decreased to ensure a variability of {\em n}-grams. The values are decreased according to the decay function in Equation~\eqref{eq:fdaequation}:

\begin{equation}\label{eq:fdaequation}
decay(f)=init(f)\frac{d^{C_L(ngr)}}{(1+C_L(ngr))^{c}} 
\end{equation}

\noindent where $C_L(ngr)$ is the count of the {\em n}-gram $ngr$ in $L$. $c$ and $d$ are parameters of FDA. By default they have a value of $0$ and $0.5$, respectively.

The $decay(ngr)$ function in Equation~\eqref{eq:fdaequation} indicates the score of the feature $ngr$ at a particular iteration, so it is dependent on the set of selected sentences $L$. 

The sentence $s$ is scored as a normalized (by length of the sentence) sum of the scores of the features. Considering the default values in Equation~\eqref{eq:fdaequation}, the resulting score function is as in Equation~\eqref{eq:fda_sentencescore}:

\begin{equation}\label{eq:fda_sentencescore}
score(s,L)=\frac{\sum_{ngr \in F_s} 0.5^{C_L(ngr)} }{\text{\# words in s}}
\end{equation}

\noindent where $F_s$ is the set of {\em n}-grams in sentence $s$.

Once the selected pool $L$ contains the desired amount of sentences, the sentences are retrieved as selected data.

\section{Experimental Setup}
\label{sec:finetun_analysis}

The data sets used in the experiments are based on the ones used in the work of \citep{biccici2013feature}:

We build German-to-English NMT model using the data provided in the WMT 2015 \citep{bojar-EtAl:2015:WMT} (4.5M sentence pairs). We consider this data set as the general-domain training data to build the non-adapted NMT (\textit{BASE}). As development data, we use 5K randomly sampled sentences from development sets of previous years.

The BASE model is adapted to two domains: news and health. Therefore we also use two test sets and two \textit{in-domain} training set (for the research question 2 and 3 explained in Section \ref{sec:finetun_RQ}):

\begin{itemize}
    \item News Domain: We use the test set provided in WMT 2015 News Translation Task, and the in-domain \textit{rapid2016}\footnote{\url{https://tilde.com/}} data set (1.3M sentence pairs) provided in  WMT 2017 News Translation\citep{bojar2017findings}.
    \item Health Domain: German-to-English parallel text from the European Medicines Agency (EMEA)\footnote{\url{http://opus.nlpl.eu/EMEA.php}} \citep{Tiedemann:RANLP5} (361K sentence pairs). For health domain test set we use the Cochrane \footnote{\url{http://www.himl.eu/test-sets}} dataset provided in WMT 2017 biomedical translation shared task \citep{yepes2017findings}.
\end{itemize}

Note that the general-domain set contains sentences from a corpus such as Europarl \citep{koehn2005europarl} which causes the domain to be closer to the news domain.

All data sets are tokenized, truecased and Byte Pair Encoding (BPE) \citep{sennrich2016neural} is applied with 89500 merge operations (the number of operations used in the work of \citet{sennrich2016neural}). The models have been built using OpenNMT-py\citep{opennmt}. We keep the default settings of OpenNMT-py: 2-layer LSTM with 500 hidden units, vocabulary size of 50000 words for each language.

We use different evaluation metrics to evaluate the performance of the models built in the experiments. These models are evaluated on the test sets using  several evaluation metrics: BLEU \citep{papineni2002bleu}, 
TER \citep{snover2006study} and METEOR \citep{banerjee2005meteor}. The scores assigned by this metrics indicate an estimation of the quality of the translation (compared to a human-translated reference). Higher scores of BLEU and METEOR indicate better translation quality. TER is an error metric, therefore lower scores indicate better performance.

In each table, scores that are better than the baseline are shown in bold. Furthermore, scores that constitute a statistically significant improvement at level p=0.01 over the baseline are marked with an asterisk. This was computed with multeval \citep{clark2011better} using Bootstrap Resampling \citep{koehn04}.

\section{Baseline Results}
\label{sec:baseline_results}

\subsection{Baseline Results with General-domain Data}

\begin{table}[!htbp]
\centering
\begin{center}
\begin{tabular}{ |p{2cm}|p{2cm}|p{2cm}|}
\hline
	&	BASE12	&	BASE13	\\
\hline	
BLEU	&	26.16	&	26.34	\\
TER 	&	54.41	&	54.41	\\
METEOR	&	30.00	&	30.09	\\
\hline	
\end{tabular}
\caption{ 
Results of the model BASE12 and BASE13 evaluated on the news test set.
 \label{table:baseline_12_13_NEWS}}
\end{center}
\end{table}

\begin{table}[!htbp]
\centering
\begin{center}
\begin{tabular}{ |p{2cm}|p{2cm}|p{2cm}|}
\hline
	&	BASE12	&	BASE13	\\
\hline	
BLEU	&	33.29	&	33.14	\\
TER 	&	46.11	&	46.79	\\
METEOR	&	34.62	&	34.57	\\
\hline	
\end{tabular}
\caption{ 
Results of the model BASE12 and BASE13 evaluated on the health test set.
\label{table:baseline_12_13_BIO}}
\end{center}
\end{table}

Table \ref{table:baseline_12_13_NEWS} presents the results evaluated with the news test set evaluated in the 12th epoch of the base model (\textit{BASE12}) and the 13th epoch (\textit{BASE13}). Similarly, Table \ref{table:baseline_12_13_BIO} presents the results evaluated with the test set in the health domain. These results help to confirm that the models trained for 12 epochs are close to convergence: In Table \ref{table:baseline_12_13_NEWS} the increment in performance from the 12th to the 13th epoch is just of 0.0018 BLEU points and in Table \ref{table:baseline_12_13_BIO} the performance is worse in the 13th epoch.


\subsection{Baseline Results With In-domain Data}

\begin{table}[!htbp]
\centering
\begin{center}
\begin{tabular}{ |p{2cm}|p{2cm}|p{2cm}|}
\hline
	&	BASE12 	&	BASE12 + rapid2016	\\
\hline	
BLEU	&	26.16	&	24.05	\\
TER 	&	54.41	&	55.86	\\
METEOR	&	30.00	&	28.74	\\
\hline	
\end{tabular}
\caption{ 
Results of the model BASE12 fine-tuned with the in-domain news set.
\label{table:baseline_12_plus_NEWS}}
\end{center}
\end{table}

\begin{table}[!htbp]
\centering
\begin{center}
\begin{tabular}{ |p{2cm}|p{2cm}|p{2cm}|}
\hline
	&	BASE12 	&	BASE12 + EMEA	\\
\hline	
BLEU	&	33.29	&	34.69	\\
TER 	&	46.11	&	44.43	\\
METEOR	&	34.62	&	34.99	\\
\hline	
\end{tabular}
\caption{ 
Results of the model BASE12 fine-tuned with the in-domain health set.
\label{table:baseline_12_plus_BIO}}
\end{center}
\end{table}

Following the work of \citet{luong2015stanford,freitag2016fast} we adapt the base system (\textit{BASE12}) by performing the 13th iteration in a different, smaller, in-domain data set. We create two new models, one adapted to the domain of news (\textit{BASE12 + rapid2016}) and another one to the health domain (\textit{BASE12 + EMEA}).

We see, in Table \ref{table:baseline_12_plus_BIO}, how using in-domain data for fine-tuning can increase the performance with more than 2 BLEU points. However, the data set chosen for performing fine-tuning is important, as in Table \ref{table:baseline_12_plus_NEWS} we see the performance of the model becomes worse after fine-tuning with the rapid2016 dataset. This also indicates that the addition of new data is not necessarily good.

\section{Main Experiments}
\label{sec:experiments}

In order to answer the questions in Section \ref{sec:finetun_RQ}, we perform three set of experiments: fine-tune the BASE12 model with a subset of the general domain data (Section \ref{sec:general_domain_models}), with a subset of in-domain data (Section \ref{sec:indomain_models}), and with a subset of data retrieved from both general domain data and in-domain data (Section \ref{sec:mixed_domain_models}).

We use the default configuration of the data selection methods. We use $d=0.5$, $c=0$ and 3-grams as features in FDA (Equation~\eqref{eq:fdaequation}).

In the INR method we also use 3-grams as $ngr$ (in Equation~\eqref{eq:infreq_ngr_recover}). In order to find a value of the threshold for the experiments, in this paper we execute several runs of INR using different values of $t$, multiplying by two in each execution (we try 10, 20, 40, 80 ...). In the experiments we use the highest value of $t$ that fulfills one of the following criteria: (i) the execution time should be under 48 hours or (ii) the number of sentences retrieved at least 500K. Accordingly, the value of $t$ in news domain is 80 (230K sentences retrieved) and in health domain 640 (275K sentences retrieved).




\subsection{Results of Models Trained in a Subset of General-Domain Data}
\label{sec:general_domain_models}

\begin{table}[!htbp]
\centering
\begin{center}
\begin{tabular}{ |p{1.2cm}|p{1.1cm}|p{1.1cm}|p{1.1cm}|p{1.1cm}|}
\hline
& \footnotesize BASE13	& \footnotesize BASE12 + TFIDF	&	\footnotesize BASE12 + INR	&	\footnotesize BASE12 + FDA \\
\hline
\multicolumn{2}{|c|}{ }	&	 \multicolumn{3}{|c|}{100K lines } \\
\hline					
BLEU	&	26.34	&	\bf26.41	&	\bf26.49	&	\bf26.49	\\
TER 	&	54.41	&	54.45	&	\bf54.19	&	\bf54.21	\\
MET.	&	30.09	&	\bf30.14	&	\bf30.21*	&	\bf30.21*	\\
\hline									
\multicolumn{2}{|c|}{ }	&	 \multicolumn{3}{|c|}{200K lines } \\
\hline									
BLEU	&	26.34	&	26.33	&	\bf26.44	&	\bf26.55*	\\
TER 	&	54.41	&	54.41	&	\bf54.35	&	\bf54.17*	\\
MET.	&	30.09	&	30.03	&	\bf30.12	&	\bf30.24*	\\
\hline									
\multicolumn{2}{|c|}{ }	&	 \multicolumn{3}{|c|}{500K lines } \\
\hline									
BLEU	&	26.34	&	\bf26.44	&	-	&	\bf26.40*	\\
TER 	&	54.41	&	\bf54.40	&	-	&	54.47	\\
MET.	&	30.09	&	\bf30.11	&	-	&	\bf30.10*	\\
\hline					
\end{tabular}
\caption{ 
Performance on the \textit{news} test for the BASE12 model, fine-tuned with subsets of the training data.
\label{table:baseline_subset_NEWS}
 }
\end{center}
\end{table}

\begin{table}[!htbp]
\centering
\begin{center}
\begin{tabular}{ |p{1.2cm}|p{1.1cm}|p{1.1cm}|p{1.1cm}|p{1.1cm}|}
\hline
& \footnotesize BASE13	& \footnotesize BASE12 + TFIDF	&	\footnotesize BASE12 + INR	&	\footnotesize BASE12 + FDA\\
\hline
\multicolumn{2}{|c|}{ }	&	 \multicolumn{3}{|c|}{100K lines } \\
\hline					
BLEU	&	33.14	&	\bf33.95*	&	\bf33.52*	&	\bf33.68*	\\
TER 	&	46.79	&	\bf45.99*	&	\bf45.92*	&	\bf45.97*	\\
MET.	&	34.57	&	\bf34.96*	&	\bf34.77	&	\bf34.71	\\
\hline									
\multicolumn{2}{|c|}{ }	&	 \multicolumn{3}{|c|}{200K lines } \\
\hline									
BLEU	&	33.14	&	\bf33.97*	&	\bf33.88*	&	\bf33.96*	\\
TER 	&	46.79	&	\bf46.03*	&	\bf45.90*	&	\bf45.64*	\\
MET.	&	34.57	&	\bf34.89*	&	\bf34.94*	&	\bf35.01*	\\
\hline									
\multicolumn{2}{|c|}{ }	&	 \multicolumn{3}{|c|}{500K lines } \\
\hline									
BLEU	&	33.14	&	\bf34.14	&	-	&	\bf33.75*	\\
TER 	&	46.79	&	\bf45.60*	&	-	&	\bf45.92*	\\
MET.	&	34.57	&	\bf34.96*	&	-	&	\bf34.92*	\\
\hline					
\end{tabular}
\caption{ 
Performance on the \textit{health} test for the BASE12 model, fine-tuned with subsets of the training data.
 \label{table:baseline_subset_BIO}
 }
\end{center}
\end{table}

In order to investigate the first question mentioned in Section \ref{sec:finetun_RQ} we select a subset of sentences of the general-domain data (the data set used to build BASE12). We extract subsets of three different sizes: 100K, 200K, and 500K lines. The only exception is the INR method which, with the established configuration, retrieves at most 230K sentences and 275K sentences using the news and health test, respectively. The BASE12 model is fine-tuned for a 13th epoch using the subset of data extracted.

In Table  \ref{table:baseline_subset_NEWS} and Table  \ref{table:baseline_subset_BIO} we show the performance of the base model in the first column (\textit{BASE13} column) and then the model in which the last epoch is fine-tuned using data selected by one of the three data selection algorithms. As we can see, fine-tuning the model with the selected data leads to improvements for most of the experiments (numbers in bold).

The vocabulary considered in the fine-tuning is the same used for building the BASE12 model. However, as BPE has been applied, this restriction is less strict. For example, in the sentence of the news test set \say{das Bildungsministerium teilte mit, etwa ein Dutzend Familien sei noch nicht zurückgekehrt.} (according to the reference, \say{the Education Ministry said about a dozen families still had not returned.}) the word \say{Bildungsministerium} (\say{Education Ministry}) would have been left out (even if in the selected data there are several occurrences) if BPE was not applied because it is infrequent in the general domain set. As in these experiments we use BPE, the adapted models achieves improvements in terms of fluency. 

The non-adapted, BASE13 model translates the above-mentioned sentence as \say{the Ministry of Education said, for example, that a dozen families did not return.}. In this sentence, the phrase \say{for example} has been added. The model adapted using TFIDF (100K lines) generates a similar sentence (i.e. \say{the Ministry of Education said, for example, that a dozen families had not returned.}), but this problem is corrected by the model adapted using INR and FDA (100K lines) as both of them generate the same translation: \say{the Ministry of Education said, about a dozen families have not returned.}. Here the phrase \say{for example} added by BASE13 model is removed.


\subsection{Results of Models Trained with a Subset of In-Domain Data}
\label{sec:indomain_models}

\begin{table}[!htbp]
\centering
\begin{center}
\begin{tabular}{ |p{1.2cm}|p{1.1cm}|p{1.1cm}|p{1.1cm}|p{1.1cm}|}
\hline
&  \footnotesize BASE12 + rapid2016	& \footnotesize BASE12 + TFIDF rapid2016	&	\footnotesize BASE12 + INR rapid2016	&	\footnotesize BASE12 + FDA rapid2016\\
\hline
\multicolumn{2}{|c|}{ }	&	 \multicolumn{3}{|c|}{100K lines } \\
\hline					
BLEU	&	24.05	&	\bf25.05*	&	\bf25.39*	&	\bf25.46*	\\
TER 	&	55.86	&	\bf55.67	&	\bf55.52*	&	\bf55.41*	\\
MET.	&	28.74	&	\bf29.07*	&	\bf29.50*	&	\bf29.49*	\\
\hline									
\multicolumn{2}{|c|}{ }	&	 \multicolumn{3}{|c|}{200K lines } \\
\hline	
BLEU	&	24.05	&	\bf24.76*	&	-	&	\bf25.12*	\\
TER 	&	55.86	&	\bf55.77	&	-	&	\bf54.76*	\\
MET.	&	28.74	&	\bf28.91	&	-	&	\bf29.54*	\\
\hline									
\multicolumn{2}{|c|}{ }	&	 \multicolumn{3}{|c|}{500K lines } \\
\hline	
BLEU	&	24.05	&	\bf24.59*	&	-	&	\bf24.75*	\\
TER 	&	55.86	&	\bf55.67	&	-	&	\bf55.10*	\\
MET.	&	28.74	&	\bf28.85	&	-	&	\bf29.33*	\\
\hline					
\end{tabular}
\caption{ 
Performance on the \textit{news} test for the BASE12 model, fine-tuned with subsets of the rapid2016 data set.
\label{table:baseline_12_subset_NEWS}
}
\end{center}
\end{table}

\begin{table}[!htbp]
\centering
\begin{center}
\begin{tabular}{ |p{1.2cm}|p{1.1cm}|p{1.1cm}|p{1.1cm}|p{1.1cm}|}
\hline
&  \footnotesize BASE12 + EMEA	& \footnotesize BASE12 + TFIDF EMEA	&	\footnotesize BASE12 + INR EMEA	&	\footnotesize BASE12 + FDA EMEA\\
\hline
\multicolumn{2}{|c|}{ }	&	 \multicolumn{3}{|c|}{100K lines } \\
\hline									
BLEU	&	34.69	&	\bf35.11	&	\bf35.22	&	\bf35.18	\\
TER 	&	44.43	&	45.09	&	\bf43.60	&	44.94	\\
MET.	&	34.99	&	\bf35.17	&	\bf35.25	&	\bf35.15	\\
\hline									
\multicolumn{2}{|c|}{ }	&	 \multicolumn{3}{|c|}{200K lines } \\
\hline
BLEU	&	34.69	&	\bf35.55	&	-	&	\bf35.11	\\
TER 	&	44.43	&	\bf44.18	&	-	&	\bf43.66	\\
MET.	&	34.99	&	\bf35.70*	&	-	&	\bf35.28	\\
\hline					
\end{tabular}
\caption{ 
Performance on the \textit{health} test for the BASE12 model, fine-tuned with subsets of the EMEA data set.
\label{table:baseline_12_subset_BIO}
}
\end{center}
\end{table}


In order to answer the second research question stated in Section \ref{sec:finetun_RQ}, we also execute the same transductive algorithms (using the same configuration) in the in-domain set (i.e. rapid2016 and EMEA). We retrieve the same amount of sentences: 100K, 200K and 500K lines for news domain; and 100K and 200K for the health domain (as EMEA only has 361K sentences). 

In Table \ref{table:baseline_12_subset_NEWS} we show in the first column, \textit{BASE12+rapid2016}, the performance of the model fine-tuned with the complete in-domain rapid2016 set (also presented in Table \ref{table:baseline_12_plus_NEWS}). The other columns contain the evaluation scores after fine-tuning BASE12 model with subsets of rapid2016. Similarly, Table \ref{table:baseline_12_subset_BIO} indicates the performance of the model fine-tuned with theEMEA dataset and different subsets (evaluated with health test). Note also that the number of sentences retrieved by INR (using the same configuration as in the previous section) is less than 200K lines, so those experiments are not executed.

Using a subset of in-domain data can improve the performance as again, most of the scores in Table \ref{table:baseline_12_subset_NEWS} and Table \ref{table:baseline_12_subset_BIO} are marked in bold. We see that the impact of the models evaluated in the news domain (Table \ref{table:baseline_12_subset_NEWS}) is higher as all experiments achieve statistically significant improvements at level p=0.01 for at least one evaluation metric. Despite that, none of the models improve the BASE13 model (column BASE13 in Table  \ref{table:baseline_12_13_NEWS}).

\subsection{Results of Models Trained with a Mixture of General-Domain and In-Domain Data}
\label{sec:mixed_domain_models}

As we have seen in previous sections, applying fine-tuning with subsets of data can perform better than using the complete dataset. In this section, we aim to explore the performance of models fine-tuned on data retrieved from a mixture of the two datasets used in previous sections: data used for building the BASE12 model, and in-domain data (rapid2016 or EMEA datasets). These experiments are particularly interesting in the case of news test because using an external dataset led to worse results.

\begin{table}[!htbp]
\centering
\begin{center}
\begin{tabular}{ |p{2cm}|p{1cm}|p{1cm}|p{1cm}|}
\hline
&  \footnotesize TFIDF	&	\footnotesize INR	&	\footnotesize FDA \\
\hline
\multicolumn{4}{|c|}{news test } \\
\hline
100K lines	&	52\%	&	89\%	&	86\%	\\
\hline							
200K lines	&	50\%	&	88\%	&	87\%	\\
\hline							
500K lines	&	46\%	&	-	&	86\%	\\
\hline
\multicolumn{4}{|c|}{health test } \\
\hline
100K lines	&	27\%	&	67\%	&	69\%	\\
\hline							
200K lines	&	29\%	&	70\%	&	71\%	\\
\hline							
500K lines	&	31\%	&	-	&	74\%	\\
\hline	
\end{tabular}
\caption{ 
Percentage of base training data lines retrieved.
\label{table:percentage_general_lines}
 }
\end{center}
\end{table}

In Table \ref{table:percentage_general_lines} we present the percentage of lines from the general domain dataset present in the selected data. We observe that in the news domain (the first subtable in Table \ref{table:percentage_general_lines}) the percentages are higher than in the health domain (the second subtable). This indicates how these transductive methods are capable of identifying better sentences. As shown in Table \ref{table:baseline_12_plus_NEWS}, the sentences from the base dataset are more useful for the news test as using the rapid2016 set for tuning the model leads to worse results.

If we perform a (column-wise) comparison of the three methods, we can observe that the INR and FDA methods retrieve a similar amount of sentences from the base set. By contrast, the TFIDF method seems to retrieve a smaller amount of sentences from the general domain data (the percentages in column TFIDF of Table \ref{table:percentage_general_lines} are much lower than the other columns).

\begin{table}[!htbp]
\centering
\begin{center}
\setlength\tabcolsep{4pt} 
\begin{tabular}{ |p{0.5cm}|p{1cm}|p{1cm}|p{1cm}|p{1.2cm}|p{1.2cm}|}
\hline
& \scriptsize BASE13 & \scriptsize BASE12 + rapid2016	& \scriptsize BASE12 + TFIDF	&	\scriptsize BASE12 + INR 	&	\scriptsize BASE12 + FDA\\
\hline
\multicolumn{3}{|c|}{ }	&	 \multicolumn{3}{|c|}{100K lines } \\
\hline	
\scriptsize BLEU	&	26.16	&	24.05	&	\bf26.42	&	\bf26.56	&	\bf26.65*	\\
\scriptsize TER 	&	54.41	&	55.86	&	54.57	&	\bf53.92*	&	\bf54.23	\\
\scriptsize MET.	&	30.09	&	28.74	&	30.06	&	\bf30.21	&	\bf30.25*	\\
\hline											
\multicolumn{3}{|c|}{ }	&	 \multicolumn{3}{|c|}{200K lines } \\
\hline											
\scriptsize BLEU	&	26.16	&	24.05	&	26.14	&	\bf26.40	&	\bf26.59	\\
\scriptsize TER 	&	54.41	&	55.86	&	54.72	&	\bf54.25	&	\bf54.22	\\
\scriptsize MET.	&	30.09	&	28.74	&	29.95	&	\bf30.13	&	\bf30.13	\\
\hline											
\multicolumn{3}{|c|}{ }	&	 \multicolumn{3}{|c|}{500K lines } \\
\hline											
\scriptsize BLEU	&	26.16	&	24.05	&	\bf26.24	&	-	&	\bf26.23	\\
\scriptsize TER 	&	54.41	&	55.86	&	54.53	&	-	&	\bf54.27	\\
\scriptsize MET.	&	30.09	&	28.74	&	29.99	&	-	&	30.02	\\
\hline					
\end{tabular}
\caption{ 
Performance on the \textit{news} test for the BASE12 model, fine-tuned with subsets of a combination of the BASE and rapid2016 data sets.
\label{table:baseline_12_mix_NEWS}
}
\end{center}
\end{table}

\begin{table}[!htbp]
\centering
\begin{center}
\setlength\tabcolsep{4pt} 
\begin{tabular}{ |p{0.5cm}|p{1cm}|p{1.1cm}|p{1.1cm}|p{1.1cm}|p{1cm}|}
\hline
& \scriptsize BASE13 & \scriptsize BASE12 + EMEA	& \scriptsize BASE12 + TFIDF	&	\scriptsize BASE12 + INR	&	\scriptsize BASE12 + FDA\\
\hline
\multicolumn{3}{|c|}{ }	&	 \multicolumn{3}{|c|}{100K lines } \\
\hline	
\scriptsize BLEU	&	33.29	&	34.69	&	34.48	&	\bf34.96	&	\bf34.89	\\
\scriptsize TER 	&	46.11	&	44.43	&	45.28	&	\bf44.68	&	\bf44.95	\\
\scriptsize MET.	&	34.62	&	34.99	&	\bf35.30	&	\bf35.35	&	\bf35.21	\\
\hline											
\multicolumn{3}{|c|}{ }	&	 \multicolumn{3}{|c|}{200K lines } \\
\hline											
\scriptsize BLEU	&	33.29	&	34.69	&	\bf35.57	&	\bf35.56	&	\bf35.59	\\
\scriptsize TER 	&	46.11	&	44.43	&	\bf44.23	&	44.59	&	45.54	\\
\scriptsize MET.	&	34.62	&	34.99	&	\bf35.59	&	\bf35.77*	&	\bf35.54	\\
\hline											
\multicolumn{3}{|c|}{ }	&	 \multicolumn{3}{|c|}{500K lines } \\
\hline											
\scriptsize BLEU	&	33.29	&	34.69	&	\bf36.79*	&	-	&	\bf35.78	\\
\scriptsize TER 	&	46.11	&	44.43	&	\bf43.30*	&	-	&	44.88	\\
\scriptsize MET.	&	34.62	&	34.99	&	\bf36.05*	&	-	&	\bf35.99	\\
\hline					
\end{tabular}
\caption{ 
Performance on the \textit{health} test for the BASE12 model, fine-tuned with subsets of a combination of the BASE and EMEA data sets.
\label{table:baseline_12_mix_BIO}
}
\end{center}
\end{table}

In Table \ref{table:baseline_12_mix_NEWS} and Table \ref{table:baseline_12_mix_BIO} we show two baselines: (i) column BASE13 shows the model built performing 13 epochs; and (ii) column BASE12+rapid2016 and BASE12+EMEA present the results observed in Table \ref{table:baseline_12_plus_NEWS} and Table \ref{table:baseline_12_plus_BIO}, respectively. In those tables we indicate in bold those scores that are better than both baselines. 

The models adapted to the news test (Table \ref{table:baseline_12_mix_NEWS}) using INR and FDA tend to perform better than both the BASE13 and the BASE12+rapid2016 models. This is especially true for smaller datasets (the adaptation with 100K lines achieves statistically significant improvements at p=0.01) but becomes closer to BASE13 when more sentences are retrieved (500K lines subtable). For the TFIDF method, despite the fact that it achieves better results than the BASE12+rapid2016 model, most of the scores are worse than the BASE13 model. As mentioned earlier, TFIDF tends to retrieve more sentences from the rapid2016 set (Table \ref{table:percentage_general_lines}), and as we saw before using more sentences from this set leads to worse performing models.

In the health domain (Table \ref{table:baseline_12_mix_BIO}), by contrast, TFIDF performs slightly better (the only experiment that achieves statistically significant improvements at p=0.01 for the three evaluation metrics).




\section{Conclusion and Future Work}
\label{sec:conclusions}

In this work, we have shown how general domain models can be adapted to a test set by fine-tuning not only to a particular domain but also to a special subset of sentences (retrieved from in-domain or out-of-domain data) that are closer to a test set and achieve better results.

We have seen that fine-tuning a model using a subset of data can achieve better performance than the model trained with the full training set. This is also applicable when using an additional set of in-domain sentences. Nonetheless, the best results are observed when augmenting the candidate sentences (i.e. combining general and in-domain sentences) as presented in Section \ref{sec:mixed_domain_models}.

FDA offers a good balance in performance and speed. INR achieve results similar to FDA, but the execution time is dependent on the configuration (i.e. value of the threshold $t$) and it may cause to exceed several hours (FDA requires less than one hour for the same execution). The configuration also restricts the amount of sentences retrieved. In the experiments performed, we retrieved no more 200K sentences to evaluate INR whereas for the other TA we could retrieve 500K parallel lines. Moreover, in this work we have used the same values of $t$ for all the experiments, which have been determined following the most restrictive assumption of not knowing the in-domain data. In the future, we want to evaluate the models fine-tuned with data retrieved from INR using different values of $t$.

TFIDF technique, although achieving comparable results, we find to be the weakest of the TA explored. The main differences with the other two is that is not a context-dependent (i.e. it does not consider the selected pool to retrieve new sentences) and in addition, each sentence is considered independently. This caused that for larger test set such \textit{news}, the improvements tend to be smaller or not to find statistically signifficant improvements at p=0.01 (e.g. tables \ref{table:baseline_subset_NEWS} and \ref{table:baseline_12_mix_NEWS}).

The experiments carried out in this paper can be further expanded using different language pairs, different domains and different selected-data sizes. Moreover, other configurations of data selection algorithms could be investigated. For example, using {\em n}-grams of higher order, executing INR with different values of $t$, in Equation~\eqref{eq:infreq_ngr_recover}, or FDA with different values of $d$ and $c$, in Equation~\eqref{eq:fdaequation} \citep{poncelasextending,poncelas2017applying}.

The techniques explored here can also be used in combination with other approaches aiming to adapt models towards a particular domain. The models presented in Section \ref{sec:mixed_domain_models} can be further expanded by adding a tag in the source sentences indicating the domain explicitly \citep{chu2017empirical,poncelas2019adapting}, using a target-side seed or using synthetic sentences \citep{chinea2017adapting,poncelas2019adaptation}.

\section*{Acknowledgements}
This research has been supported by the ADAPT Centre for Digital Content Technology which is funded under the SFI Research Centres Programme (Grant 13/RC/2106) and is co-funded under the European Regional Development Fund.

\noindent 
\includegraphics[width=1cm]{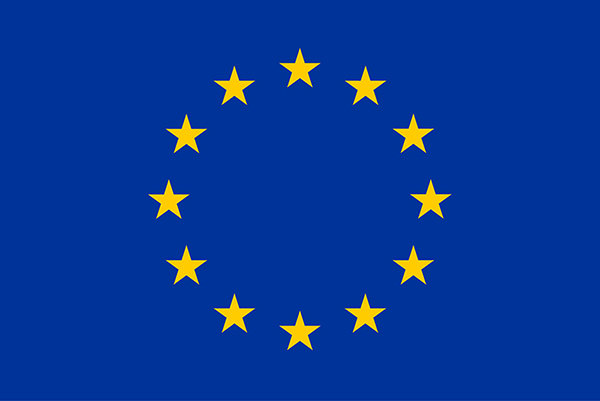}
This work has also received funding from the European Union’s Horizon 2020 research and innovation programme under the Marie Skłodowska-Curie grant agreement No 713567.

\bibliographystyle{plain}
\bibliography{bibl}

\begin{thebibliography}{10}

\bibitem{axelrod2011domain}
Amittai Axelrod, Xiaodong He, and Jianfeng Gao.
\newblock Domain adaptation via pseudo in-domain data selection.
\newblock In {\em Proceedings of the 2011 Conference on Empirical Methods in
  Natural Language Processing}, pages 355--362, Edinburgh, Scotland, UK., 2011.

\bibitem{banerjee2005meteor}
Satanjeev Banerjee and Alon Lavie.
\newblock Meteor: An automatic metric for {MT} evaluation with improved
  correlation with human judgments.
\newblock In {\em Proceedings of the {ACL} workshop on intrinsic and extrinsic
  evaluation measures for machine translation and/or summarization}, pages
  65--72, Ann Arbor, Michigan, 2005.

\bibitem{biccici2013feature}
Ergun Bi{\c{c}}ici.
\newblock Feature decay algorithms for fast deployment of accurate statistical
  machine translation systems.
\newblock In {\em Proceedings of the Eighth Workshop on Statistical Machine
  Translation}, pages 78--84, Sofia, Bulgaria, August 2013.

\bibitem{biccici2011instance}
Ergun Bi{\c{c}}ici and Deniz Yuret.
\newblock Instance selection for machine translation using feature decay
  algorithms.
\newblock In {\em Proceedings of the Sixth Workshop on Statistical Machine
  Translation}, pages 272--283, Edinburgh, Scotland, 2011.

\bibitem{bojar2017findings}
Ond{\v{r}}ej Bojar, Rajen Chatterjee, Christian Federmann, Yvette Graham, Barry
  Haddow, Shujian Huang, Matthias Huck, Philipp Koehn, Qun Liu, Varvara
  Logacheva, et~al.
\newblock Findings of the 2017 conference on machine translation (wmt17).
\newblock In {\em Proceedings of the Second Conference on Machine Translation},
  pages 169--214, Copenhagen, Denmark, 2017.

\bibitem{bojar-EtAl:2015:WMT}
Ond\v{r}ej Bojar, Rajen Chatterjee, Christian Federmann, Barry Haddow, Matthias
  Huck, Chris Hokamp, Philipp Koehn, Varvara Logacheva, Christof Monz, Matteo
  Negri, Matt Post, Carolina Scarton, Lucia Specia, and Marco Turchi.
\newblock {Findings of the 2015 Workshop on Statistical Machine Translation}.
\newblock In {\em {Proceedings of the Tenth Workshop on Statistical Machine
  Translation}}, pages 1--46, Lisboa, Portugal, 2015.

\bibitem{chinea2017adapting}
Mara Chinea-Rios, Alvaro Peris, and Francisco Casacuberta.
\newblock Adapting neural machine translation with parallel synthetic data.
\newblock In {\em Proceedings of the Second Conference on Machine Translation},
  pages 138--147, Copenhagen, Denmark, 2017.

\bibitem{chu2017empirical}
Chenhui Chu, Raj Dabre, and Sadao Kurohashi.
\newblock An empirical comparison of domain adaptation methods for neural
  machine translation.
\newblock In {\em Proceedings of the 55th Annual Meeting of the Association for
  Computational Linguistics (Volume 2: Short Papers)}, volume~2, pages
  385--391, Vancouver, Canada, 2017.

\bibitem{chu2018survey}
Chenhui Chu and Rui Wang.
\newblock A survey of domain adaptation for neural machine translation.
\newblock {\em arXiv preprint arXiv:1806.00258}, 2018.

\bibitem{clark2011better}
Jonathan~H. Clark, Chris Dyer, Alon Lavie, and Noah~A. Smith.
\newblock Better hypothesis testing for statistical machine translation:
  Controlling for optimizer instability.
\newblock In {\em Proceedings of the 49th Annual Meeting of the Association for
  Computational Linguistics: Human Language Technologies (Volume 2: Short
  Papers)}, page 176–181, Portland, Oregon, 2011.

\bibitem{eetemadi2015survey}
Sauleh Eetemadi, William Lewis, Kristina Toutanova, and Hayder Radha.
\newblock Survey of data-selection methods in statistical machine translation.
\newblock {\em Machine Translation}, 29(3-4):189--223, 2015.

\bibitem{freitag2016fast}
Markus Freitag and Yaser Al-Onaizan.
\newblock Fast domain adaptation for neural machine translation.
\newblock {\em arXiv preprint arXiv:1612.06897}, 2016.

\bibitem{gasco2012does}
Guillem Gasc{\'o}, Martha-Alicia Rocha, Germ{\'a}n Sanchis-Trilles, Jes{\'u}s
  Andr{\'e}s-Ferrer, and Francisco Casacuberta.
\newblock Does more data always yield better translations?
\newblock In {\em Proceedings of the 13th Conference of the European Chapter of
  the Association for Computational Linguistics}, pages 152--161, Avignon,
  France, 2012.

\bibitem{hildebrand2005adaptation}
Almut~Silja Hildebrand, Matthias Eck, Stephan Vogel, and Alex Waibel.
\newblock Adaptation of the translation model for statistical machine
  translation based on information retrieval.
\newblock In {\em Proceedings of the 10th Annual Conference of the European
  Association for Machine Translation}, pages 133--142, Budapest, Hungary,
  2005.

\bibitem{opennmt}
Guillaume Klein, Yoon Kim, Yuntian Deng, Jean Senellart, and Alexander~M. Rush.
\newblock Opennmt: Open-source toolkit for neural machine translation.
\newblock In {\em Proceedings of the 55th Annual Meeting of the Association for
  Computational Linguistics-System Demonstrations}, pages 67--72, Vancouver,
  Canada, 2017.

\bibitem{koehn04}
Philipp Koehn.
\newblock Statistical significance tests for machine translation evaluation.
\newblock In {\em Proceedings of the 2004 Conference on Empirical Methods in
  Natural Language Processing}, pages 388--395, Barcelona, Spain, 2004.

\bibitem{koehn2005europarl}
Philipp Koehn.
\newblock Europarl: A parallel corpus for statistical machine translation.
\newblock {\em Machine Translation Summit, 2005}, pages 79--86, 2005.

\bibitem{levenshtein1966binary}
Vladimir Levenshtein.
\newblock Binary codes capable of correcting deletions, insertions and
  reversals.
\newblock In {\em Soviet Physics Doklady}, pages 707--710, 1966.

\bibitem{li2018one}
Xiaoqing Li, Jiajun Zhang, and Chengqing Zong.
\newblock One sentence one model for neural machine translation.
\newblock In {\em Proceedings of the Eleventh International Conference on
  Language Resources and Evaluation (LREC 2018)}, pages 910--917, Miyazaki,
  Japan, 2018.

\bibitem{liu2012locally}
Lemao Liu, Hailong Cao, Taro Watanabe, Tiejun Zhao, Mo~Yu, and Conghui Zhu.
\newblock Locally training the log-linear model for smt.
\newblock In {\em Proceedings of the 2012 Joint Conference on Empirical Methods
  in Natural Language Processing and Computational Natural Language Learning},
  pages 402--411, Jeju, Korea, 2012.

\bibitem{luong2015stanford}
Minh-Thang Luong and Christopher~D Manning.
\newblock Stanford neural machine translation systems for spoken language
  domains.
\newblock In {\em Proceedings of the International Workshop on Spoken Language
  Translation}, pages 76--79, Da Nang, Vietnam, 2015.

\bibitem{mikolov2013distributed}
Tomas Mikolov, Ilya Sutskever, Kai Chen, Greg~S Corrado, and Jeff Dean.
\newblock Distributed representations of words and phrases and their
  compositionality.
\newblock In {\em Advances in neural information processing systems}, pages
  3111--3119, 2013.

\bibitem{papineni2002bleu}
Kishore Papineni, Salim Roukos, Todd Ward, and Wei-Jing Zhu.
\newblock Bleu: a method for automatic evaluation of machine translation.
\newblock In {\em Proceedings of 40th Annual Meeting of the Association for
  Computational Linguistics}, pages 311--318, Philadelphia, Pennsylvania, USA,
  July 2002.

\bibitem{parcheta2018data}
Zuzanna Parcheta, Germ{\'a}n Sanchis-Trilles, and Francisco Casacuberta.
\newblock Data selection for nmt using infrequent n-gram recovery.
\newblock In {\em Proceedings of the 21st Annual Conference of the European
  Association for Machine Translation}, page 219–227, Alacant, Spain, 2018.

\bibitem{poncelas2017applying}
Alberto Poncelas, Gideon~Maillette de~Buy~Wenniger, and Andy Way.
\newblock Applying n-gram alignment entropy to improve feature decay
  algorithms.
\newblock {\em The Prague Bulletin of Mathematical Linguistics},
  108(1):245--256, 2017.

\bibitem{poncelas2018data}
Alberto Poncelas, Gideon~Maillette de~Buy~Wenniger, and Andy Way.
\newblock Data selection with feature decay algorithms using an approximated
  target side.
\newblock In {\em 15th International Workshop on Spoken Language Translation
  (IWSLT 2018)}, pages 173--180, Bruges, Belgium, 2018.

\bibitem{poncelas2018feature}
Alberto Poncelas, Gideon~Maillette de~Buy~Wenniger, and Andy Way.
\newblock Feature decay algorithms for neural machine translation.
\newblock In {\em Proceedings of the 21st Annual Conference of the European
  Association for Machine Translation}, pages 239--248, Alacant, Spain, 2018.

\bibitem{poncelas2019adaptation}
Alberto Poncelas, Gideon~Maillette de~Buy~Wenniger, and Andy Way.
\newblock Adaptation of machine translation models with back-translated data
  using transductive data selection methods.
\newblock In {\em 20th International Conference on Computational Linguistics
  and Intelligent Text Processing}, La Rochelle, France, 2019.

\bibitem{poncelas2019adapting}
Alberto Poncelas, Kepa Sarasola, Meghan Dowling, Andy Way, Gorka Labaka, and
  Iñaki Alegria.
\newblock {Adapting NMT to caption translation in Wikimedia Commons for
  low-resource languages}.
\newblock In {\em 35th International Conference of the Spanish Society for
  Natural Language Processing (SEPLN 2019)}, Bilbao, Spain, 2019.

\bibitem{poncelas2018adapt}
Alberto Poncelas, Andy Way, and Kepa Sarasola.
\newblock {The ADAPT System Description for the IWSLT 2018 Basque to English
  Translation Task}.
\newblock In {\em International Workshop on Spoken Language Translation}, pages
  72--82, Bruges, Belgium, 2018.

\bibitem{poncelasextending}
Alberto Poncelas, Andy Way, and Antonio Toral.
\newblock Extending feature decay algorithms using alignment entropy.
\newblock In {\em International Workshop on Future and Emerging Trends in
  Language Technology}, pages 170--182, Seville, Spain, 2016.

\bibitem{tfidf1973}
Gerard Salton and Chung-Shu Yang.
\newblock On the specification of term values in automatic indexing.
\newblock {\em Journal of documentation}, 29(4):351--372, 1973.

\bibitem{sennrich2016neural}
Rico Sennrich, Barry Haddow, and Alexandra Birch.
\newblock Neural machine translation of rare words with subword units.
\newblock In {\em Proceedings of the 54th Annual Meeting of the Association for
  Computational Linguistics (Volume 1: Long Papers)}, volume~1, pages
  1715--1725, Berlin, Germany, 2016.

\bibitem{silva2018extracting}
Catarina~Cruz Silva, Chao-Hong Liu, Alberto Poncelas, and Andy Way.
\newblock Extracting in-domain training corpora for neural machine translation
  using data selection methods.
\newblock In {\em Proceedings of the Third Conference on Machine Translation:
  Research Papers}, pages 224--231, Brussels, Belgium, 2018.

\bibitem{snover2006study}
Matthew Snover, Bonnie Dorr, Richard Schwartz, Linnea Micciulla, and John
  Makhoul.
\newblock A study of translation edit rate with targeted human annotation.
\newblock In {\em Proceedings of the 7th Conference of the Association for
  Machine Translation in the Americas}, pages 223--231, Cambridge,
  Massachusetts, USA, 2006.

\bibitem{Tiedemann:RANLP5}
J\"org Tiedemann.
\newblock News from {OPUS} - {A} collection of multilingual parallel corpora
  with tools and interfaces.
\newblock In N.~Nicolov, K.~Bontcheva, G.~Angelova, and R.~Mitkov, editors,
  {\em Recent Advances in Natural Language Processing}, volume~V, pages
  237--248. John Benjamins, Amsterdam/Philadelphia, Borovets, Bulgaria, 2009.

\bibitem{utiyama2009two}
Masao Utiyama, Hirofumi Yamamoto, and Eiichiro Sumita.
\newblock Two methods for stabilizing mert: Nict at iwslt 2009.
\newblock In {\em International Workshop on Spoken Language Translation (IWSLT
  2009)}, pages 79--82, Tokyo, Japan, 2009.

\bibitem{van2017dynamic}
Marlies van~der Wees, Arianna Bisazza, and Christof Monz.
\newblock Dynamic data selection for neural machine translation.
\newblock In {\em Proceedings of the 2017 Conference on Empirical Methods in
  Natural Language Processing}, pages 1400--1410, Copenhagen, Denmark, 2017.

\bibitem{Vapnik1998}
Vladimir~N. Vapnik.
\newblock {\em Statistical Learning Theory}.
\newblock Wiley-Interscience, 1998.

\bibitem{yepes2017findings}
Antonio~Jimeno Yepes, Aur{\'e}lie N{\'e}v{\'e}ol, Mariana Neves, Karin
  Verspoor, Ondrej Bojar, Arthur Boyer, Cristian Grozea, Barry Haddow,
  Madeleine Kittner, Yvonne Lichtblau, et~al.
\newblock Findings of the wmt 2017 biomedical translation shared task.
\newblock In {\em Proceedings of the Second Conference on Machine Translation},
  pages 234--247, 2017.

\end{thebibliography}

\end{document}